\documentclass[11pt]{article}

\usepackage[preprint]{acl}
\setcitestyle{numbers}

\usepackage{times}
\usepackage{latexsym}

\usepackage[T1]{fontenc}
\usepackage[utf8]{inputenc}
\usepackage{microtype}
\usepackage{inconsolata}
\usepackage{graphicx}
\usepackage{booktabs}
\usepackage{subcaption}

\title{Robust Marathi Sentiment Analysis using Domain-Adaptive Pre-training and Data Augmentation Approaches}
\author{
Tanvi Somani$^{1,3}$, Aryan Babare$^{1,3}$, Samyak Bora$^{1,3}$, Raviraj Joshi$^{2,3}$, Abhinay Dhamankar$^{1}$ \\
$^{1}$Pune Institute of Computer Technology (PICT), Pune, Maharashtra, India \\
$^{2}$Indian Institute of Technology Madras, Chennai, Tamil Nadu, India \\
$^{3}$L3Cube Labs, Pune \\
\texttt{\{tanvisomani0216, aryanbabare1, samyakkiranbora\}@gmail.com} \\
\texttt{agdhamankar@pict.edu} \\
\texttt{ravirajoshi@gmail.com}
}

\begin{document}
\maketitle

\begin{abstract}
For languages with limited resources, Sentiment Analysis remains a challenging problem due to limited annotated data and strong domain variability across textual sources. Marathi, despite being widely spoken, lacks extensive multi-domain sentiment analysis studies. Recent efforts by L3Cube have introduced multi-domain datasets in Marathi, enabling a better study of domain-aware sentiment classification.
This study uses the pretrained MahaBERT model to improve sentiment classification for Marathi across different domains. Models are trained on individual domain datasets and domain-adaptive pretraining is applied using masked language modeling to help the model better understand domain-specific language. In addition, back-translation is used as a method of data augmentation to increase the diversity of training data. Results show that these methods lead to consistent improvements in accuracy across all domains. The improvements range from a minimum of about 0.12\% to a maximum of about 1.74\%, depending on the dataset.
These results demonstrate that domain-adaptive pretraining and data augmentation are effective strategies for improving sentiment analysis models. This helps achieve more robust performance across different domains and makes the approach more suitable for low-resource languages like Marathi.
\end{abstract}

\section{Introduction}

The interest in sentiment analysis in resource-constrained Indic languages has increased in recent times due to the rapid increase in user-generated content on digital platforms. However, the development of effective sentiment classification models for these languages has been challenging due to limited resources, linguistic differences, and large variability across domains. Although Marathi is a widely spoken language, there is limited availability of large-scale resources for sentiment analysis, which makes achieving strong performance across varied domains difficult.

Recently, a new multi-domain Marathi sentiment analysis dataset, known as MahaSent-MD, has been introduced, which has 60K labeled sentences for positive, negative, and neutral sentiment \cite{pingle2023l3cube, kulkarni2021l3cubemahasent}. The dataset spans four distinct domains with approximately equal sample distribution across each domain, making it well-suited for systematic experimentation and performance analysis across multiple domains. The availability of this resource allows for the evaluation of sentiment analysis models across diverse domain settings. Experimental results have shown that the monolingual MahaBERT model \cite{joshi2022l3cube} has superior performance compared to multilingual transformer models across all domains.

The importance of domain diversity in sentiment classification performance was also discussed in previous studies \cite{roccabruna2022multi,mamta2022hindimd}. Mamta et al. introduced the HindiMD corpus for the purpose of testing the effectiveness of domain-aware approaches in the context of the Hindi language \cite{mamta2022hindimd}. Roccabruna et al. also found that exposing models to multiple domains helps them learn more robust representations \cite{roccabruna2022multi}.

Recently, domain-adaptive pretraining was found to be effective in further improving the performance \cite{rietzler2020adapt}. Rietzler et al. presented a method in which the masked language modeling approach is used to adapt the BERT model \cite{devlin2019bert} to domain-specific data before fine-tuning.

In addition to this, data augmentation methods including back-translation have been demonstrated to work well for low-resource languages by introducing paraphrase diversity into the training data \cite{sugiyama2019data,pingle2023robust}.

Overall, the above studies emphasize the significance of domain-aware training, adaptive pre-training, and data augmentation for building effective sentiment analysis tools for low-resource languages such as Marathi.

The following are this work's primary contributions:
\begin{itemize}
    \item Investigated domain-specific sentiment classification for Marathi using the pretrained MahaBERT model and evaluated performance across individual domains of the MahaSent-MD dataset.
    \item Applied domain-adaptive pretraining via masked language modeling and back-translation-based data augmentation to improve contextual understanding and increase training data diversity.
    \item Demonstrated consistent and reliable performance improvements across all domains, validating the effectiveness of adaptive training strategies for low-resource Marathi sentiment analysis.
\end{itemize}

\section{Related Work}

In recent years, sentiment analysis in low-resource languages has received a lot of interest due to the scarcity of annotated data and the presence of strong domain variability across textual sources. One of the key directions in this area is the development of domain-diverse datasets that enable robust evaluation of sentiment classification models. The MahaSent-MD dataset provides a comprehensive multi-domain Marathi benchmark, covering multiple types of text such as reviews and social media content \cite{pingle2023l3cube}. Similarly, the HindiMD dataset further emphasizes the importance of domain diversity for improving sentiment analysis in low-resource settings \cite{mamta2022hindimd}.

These datasets highlight that linguistic patterns and sentiment expressions differ significantly across domains, making domain-aware modeling a challenging yet essential task.

With the rise of deep learning, transformer-based models have become the dominant approach for sentiment analysis. BERT, introduced by Devlin et al.\ \cite{devlin2019bert}, uses masked language modeling to learn deep contextual representations, achieving state-of-the-art results across many NLP tasks. Later research confirms that these models clearly outperform traditional machine learning methods, especially in low-resource languages \cite{aliyu2024comparative}, though the quality and diversity of training data still heavily influences their effectiveness. While multilingual models offer broad language coverage, monolingual models tend to perform better for specific languages due to their ability to capture fine-grained linguistic features \cite{aliyu2024sentiment}.

In order to overcome this restriction, monolingual transformer models such as MahaBERT have been developed specifically for Marathi. These models are trained on large-scale Marathi corpora and are shown to outperform multilingual alternatives in downstream tasks such as sentiment classification, text classification, and named entity recognition \cite{joshi2022l3cube}. Additionally, the L3Cube-MahaNLP initiative provides a suite of Marathi NLP resources, including datasets and pretrained models, which further support research in this domain \cite{joshi2022mahanp}. The strong performance of these monolingual models motivates the selection of MahaBERT as the base model for domain-aware sentiment classification.

Domain adaptation has been widely studied to boost model performance on domain-specific data. A common approach combines domain-adaptive pretraining using masked language modeling (MLM) with task-specific fine-tuning. Rietzler et al.\ \cite{rietzler2020adapt} propose a two-step framework where BERT is first adapted to domain-specific data and then fine-tuned for sentiment classification, yielding better domain performance. Kotelnikova et al.\ \cite{kotelnikova2023cross} similarly show that even a small amount of in-domain fine-tuning data can significantly improve results. Exposing models to diverse linguistic patterns through multi-domain training further enhances generalization and reduces domain-specific bias \cite{roccabruna2022multi,santos2021multi}, underscoring that combining domain adaptation with varied training data is key to robust sentiment classification.

Data augmentation is another key strategy for improving performance in low-resource settings. Back-translation, one of the most widely used techniques, generates paraphrases by translating sentences into another language and back, increasing data diversity \cite{sugiyama2019data}. Simpler methods like EDA apply synonym replacement, random insertion, and deletion to expand training datasets \cite{wei2019eda}. More advanced approaches such as SSMBA leverage self-supervised learning to produce meaningful augmented samples that enhance robustness \cite{ng2020ssmba}. Augmentation methods applied specifically to Marathi sentiment analysis further confirm that greater data diversity leads to consistent performance gains \cite{pingle2023robust}.

In addition to these approaches, alternative techniques such as Siamese networks have been explored to learn sentiment representations for resource-poor languages by leveraging similarity-based learning \cite{choudhary2018emotions}. Comparative studies of transformer models also show that model performance varies depending on domain characteristics and training strategies, reinforcing the need for domain-aware approaches \cite{aliyu2024comparative}. Recent surveys summarize that combining multiple strategies---including domain adaptation, multi-domain training, and data augmentation---leads to more robust and generalized sentiment classification systems \cite{aliyu2024sentiment}.

Sentiment analysis in low-resource languages faces challenges such as class imbalance, noisy data, and domain variability, which limit model performance. While transformer-based models provide strong contextual representations, their effectiveness depends on training data diversity and domain alignment. Recent studies show that multi-domain training and data augmentation improve robustness by increasing data diversity, while domain-adaptive pretraining enhances domain-specific performance \cite{aliyu2024comparative,aliyu2024sentiment}. Although cross-lingual approaches offer some benefits, they often fail to capture language-specific nuances, highlighting the importance of monolingual models.

Overall, existing research indicates that no single approach is adequate to deal with the challenges of sentiment analysis in low-resource languages. Instead, integrating domain-diverse datasets, monolingual transformer models, domain-adaptive pretraining, and data augmentation techniques provides the most effective solution for improving domain-specific performance.

\section{Methodology}

\begin{figure*}[t]
  \centering
  \includegraphics[width=0.85\textwidth]{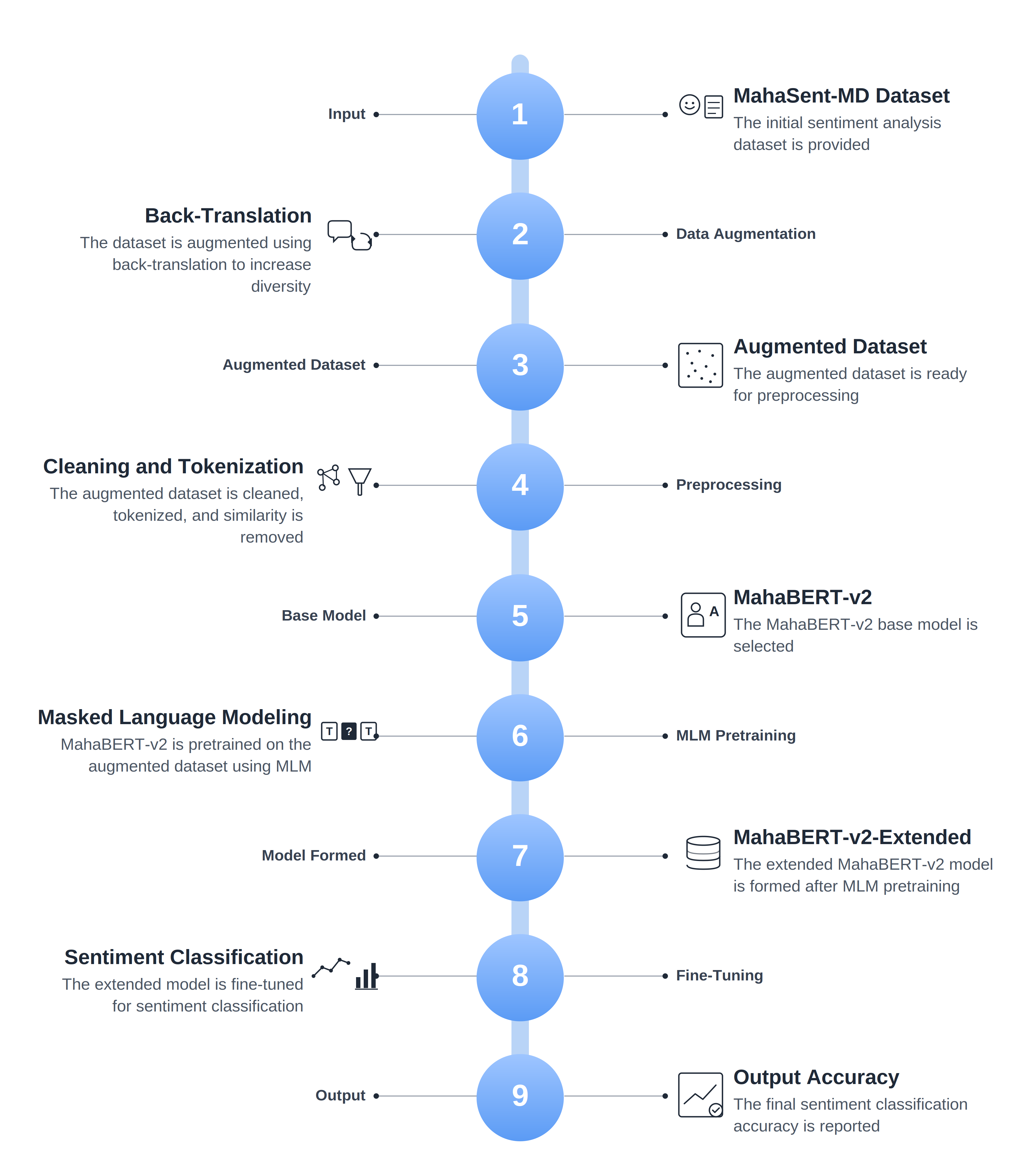}
  \caption{Architecture for Marathi sentiment classification using MahaBERT.}
  \label{fig:architecture}
\end{figure*}

The general design of the suggested Marathi sentiment classification system is depicted in Figure~\ref{fig:architecture}. The proposed framework is designed to effectively address domain-specific sentiment classification by integrating data augmentation, domain-adaptive pretraining, and supervised fine-tuning within a unified transformer-based pipeline. The architecture is built on the MahaBERT model \cite{joshi2022l3cube}, a monolingual BERT-based transformer pretrained on a large Marathi corpus, which enables better understanding of language-specific features and contextual relationships.

\subsection{Datasets}

In order to assess the suggested model's performance, experiments are conducted using the MahaSent-MD dataset \cite{pingle2023l3cube}, a multi-domain sentiment dataset in Marathi designed for sentiment classification tasks. It consists of four distinct domains:

\begin{itemize}
    \item \textbf{MahaSent-MR} (Movie Reviews)
    \item \textbf{MahaSent-PT} (Political Tweets)
    \item \textbf{MahaSent-GT} (Generic Tweets)
    \item \textbf{MahaSent-ST} (Subtitles)
\end{itemize}

These domains capture a variety of linguistic variations, writing styles, and contextual differences, reflecting real-world diversity in Marathi text data. One of three sentiment labels---positive, negative, or neutral---is applied to each instance in the collection. Importantly, the dataset maintains approximately equal sample distribution across all four domains, which ensures balanced representation and makes systematic experimentation and comparative analysis across domains feasible.

The variation in domain characteristics introduces challenges such as class imbalance and domain-specific language patterns, which are common in real-world applications. Additionally, the dataset includes both formal text (such as movie reviews and subtitles) and informal text (such as tweets), allowing the model to learn diverse linguistic features including slang, abbreviations, and contextual expressions.

Overall, the MahaSent-MD dataset provides a strong benchmark for domain-specific sentiment classification and enables effective evaluation of the impact of techniques such as domain-adaptive pretraining and data augmentation in low-resource language settings.

\subsection{Data Augmentation}

\begin{figure*}[t]
\centering
\begin{subfigure}[b]{0.48\textwidth}
    \centering
    \includegraphics[width=\textwidth]{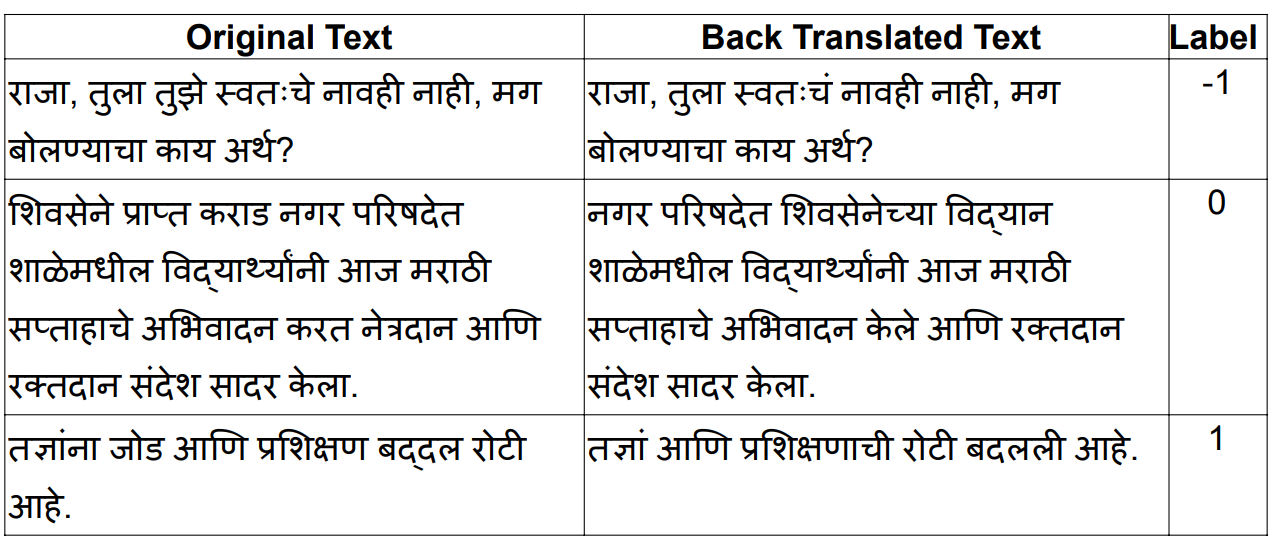}
    \caption{Augmented PT dataset}
\end{subfigure}
\hfill
\begin{subfigure}[b]{0.48\textwidth}
    \centering
    \includegraphics[width=\textwidth]{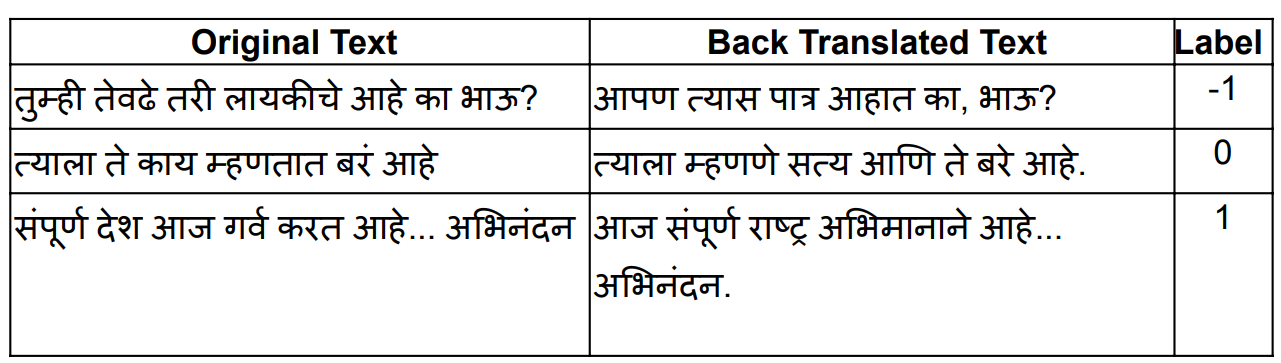}
    \caption{Augmented GT dataset}
\end{subfigure}

\vspace{0.3cm}

\begin{subfigure}[b]{0.48\textwidth}
    \centering
    \includegraphics[width=\textwidth]{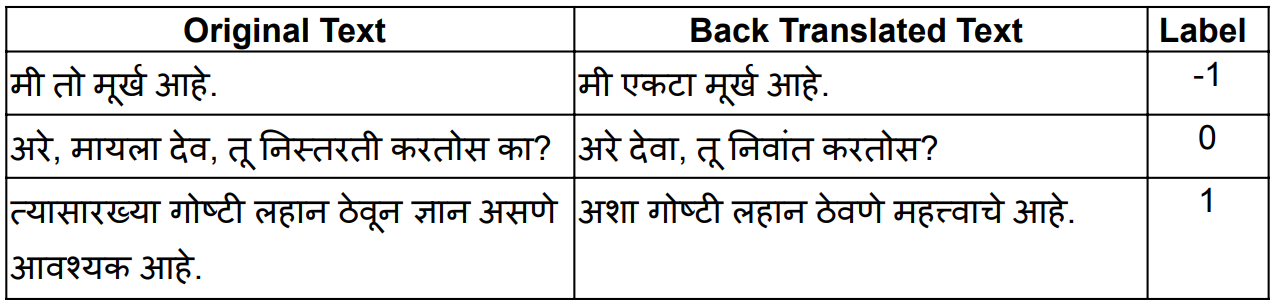}
    \caption{Augmented ST dataset}
\end{subfigure}
\hfill
\begin{subfigure}[b]{0.48\textwidth}
    \centering
    \includegraphics[width=\textwidth]{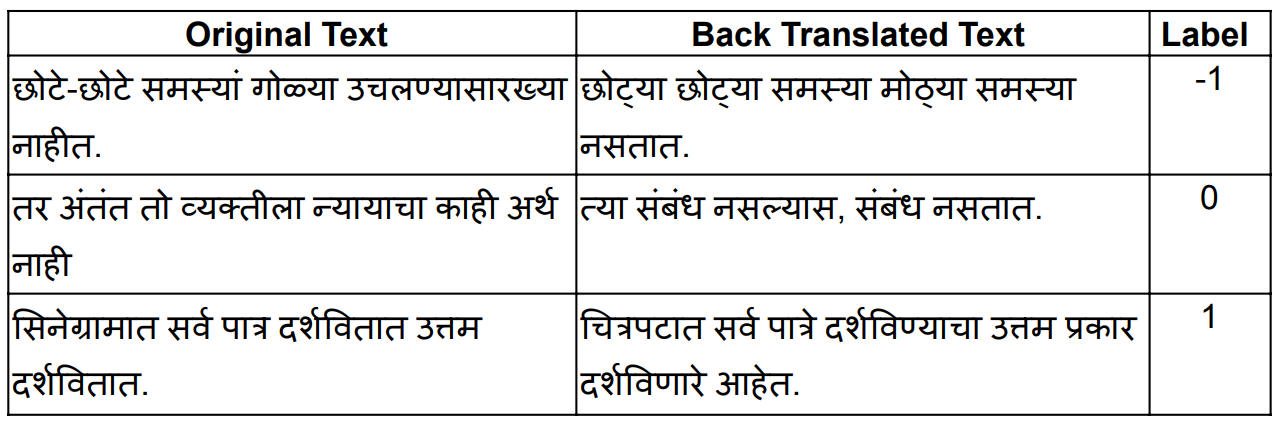}
    \caption{Augmented MR dataset}
\end{subfigure}

\caption{Comparison of Original and Augmented Data across domains.}
\label{fig:dataset}
\end{figure*}

The process begins with the MahaSent-MD dataset, which consists of multiple domains with diverse linguistic characteristics. Since low-resource datasets often suffer from limited data and domain variability, back-translation \cite{sugiyama2019data} is used as a data augmentation method. In this step, Marathi sentences are translated into another language and then back into Marathi to create paraphrases of the original text. This increases the dataset's diversity and helps the model learn more varied linguistic expressions within each domain \cite{pingle2023robust}. Figure~\ref{fig:dataset} shows a comparison of the original and augmented data across all four domains.

Following augmentation, the dataset undergoes preprocessing, which includes text cleaning, normalization, and tokenization using the MahaBERT tokenizer. This stage ensures that the input content is formatted appropriately for the transformer model while preserving important contextual information and removing noise.

\subsection{Preprocessing}

The dataset undergoes preprocessing, including text cleaning, removal of empty samples, and normalization. The processed text is then tokenized using the MahaBERT tokenizer \cite{joshi2022l3cube}. Each input sentence is converted into token IDs and attention masks, and truncated or padded to a maximum sequence length of 128 tokens.

\subsection{Domain-Adaptive Pretraining (MLM)}

The central element of the architecture is the MahaBERT model. In the first stage, domain-adaptive pretraining is performed using Masked Language Modeling (MLM) \cite{devlin2019bert,rietzler2020adapt}. During this process, the model is trained to predict a portion of the randomly masked tokens in each input sequence using domain-specific text. This self-supervised learning step allows the model to capture domain-specific vocabulary, contextual patterns, and linguistic variations present in the augmented dataset.

After MLM pretraining, an adapted version of the MahaBERT model is obtained, which incorporates improved domain-specific contextual representations. In the second stage, this adapted model is fine-tuned for the sentiment classification task. A classification head consisting of a fully connected layer and a softmax activation function is placed on top of the transformer. The model predicts one of three sentiment classes: \textbf{positive, negative, or neutral}.

\subsection{Fine-Tuning for Sentiment Classification}

During fine-tuning, the \textbf{AdamW optimizer} is used to optimize the model parameters, with \textbf{cross-entropy} loss serving as the objective function. Early stopping based on validation loss is employed to guarantee stable convergence, and a \textbf{dropout layer} is used to reduce overfitting and enhance generalization.

The model is trained for multiple epochs with a carefully selected learning rate to balance convergence speed and performance. Additionally, evaluation is performed at each epoch to monitor model performance and prevent degradation. The key hyperparameters used during both the MLM pretraining and fine-tuning stages are summarized in Table~\ref{tab:hyperparams}.

\subsection{Evaluation Metrics}

There are several parameters, including accuracy, precision, recall, and F1-score, that are used to evaluate the performance of the model. Accuracy is taken as the primary parameter for evaluating the performance of the model, whereas the other parameters provide a more detailed analysis of the performance of the model.

An overview of the training process, along with hyperparameters, is provided in Table~\ref{tab:hyperparams}. This approach helps the model gain knowledge about general language as well as domain-specific features, resulting in improved performance in sentiment classification within respective domains.

\begin{table}[t]
  \centering
  \small
  \begin{tabular}{ll}
    \toprule
    \textbf{Parameter} & \textbf{Value} \\
    \midrule
    \multicolumn{2}{l}{\textit{Model Configuration}} \\
    Pretrained Model & MahaBERT (marathi-bert-v2) \\
    Embedding Size & 768 \\
    Max Sequence Length & 128 \\
    \midrule
    \multicolumn{2}{l}{\textit{MLM Pretraining}} \\
    Task & Masked Language Modeling \\
    Masking Probability & 15\% \\
    Batch Size & 16 \\
    Learning Rate & 5e-5 \\
    Epochs & 3 \\
    Optimizer & AdamW (fused) \\
    Evaluation Strategy & Steps \\
    FP16 & True \\
    \midrule
    \multicolumn{2}{l}{\textit{Fine-Tuning (Sentiment Classification)}} \\
    Batch Size & 16 \\
    Learning Rate & 2e-5 \\
    Epochs & 3 \\
    Optimizer & AdamW \\
    Evaluation Strategy & Epoch \\
    Dropout & 0.1 \\
    Loss Function & Cross Entropy Loss \\
    FP16 & False \\
    \midrule
    \multicolumn{2}{l}{\textit{Training Settings}} \\
    Random Seed & 42 \\
    Evaluation Metrics & Accuracy, Precision, Recall, F1 \\
    \bottomrule
  \end{tabular}
  \caption{Training Hyperparameters for MLM Pretraining and Fine-Tuning.}
  \label{tab:hyperparams}
\end{table}

\section{Results and Discussion}

This section presents the experimental results and examines the effectiveness of domain-adaptive pretraining and data augmentation in improving sentiment classification performance. The model is evaluated across all domains of the MahaSent-MD dataset, and its performance is compared against a baseline model to quantify the contribution of each proposed technique.

\begin{table}[t]
\centering
\small
\setlength{\tabcolsep}{6pt}

\begin{tabular}{lcc}
\toprule
\textbf{Dataset} & \textbf{Baseline (\%)} & \textbf{Improved (\%)} \\
\midrule
MR (Movie Reviews)    & 78.90 & 79.88 \\
PT (Political Tweets) & 84.90 & 85.02 \\
GT (Generic Tweets)   & 78.80 & 79.20 \\
ST (Subtitles)        & 79.06 & 80.80 \\
\bottomrule
\end{tabular}

\caption{Baseline vs Improved Accuracy Comparison.}
\label{tab:accuracy}
\end{table}

\begin{table}[t]
\centering
\small
\setlength{\tabcolsep}{9pt}

\begin{tabular}{lcc}
\toprule
\textbf{Dataset} & \textbf{Baseline (\%)} & \textbf{Gain (\%)} \\
\midrule
MR (Movie Reviews)    & 78.90 & +0.98 \\
PT (Political Tweets) & 84.90 & +0.12 \\
GT (Generic Tweets)   & 78.80 & +0.40 \\
ST (Subtitles)        & 79.06 & +1.74 \\
\bottomrule
\end{tabular}

\caption{Accuracy Gain from Proposed Approach.}
\label{tab:gain}
\end{table}

As shown in Tables~\ref{tab:accuracy} and~\ref{tab:gain}, the proposed model consistently outperforms the baseline across all domains. These improvements confirm that domain-adaptive pretraining and data augmentation are effective strategies for improving the model's ability to capture domain-specific patterns and produce more reliable predictions.

Among all domains, the highest improvement is observed in the ST (Subtitles) dataset, with a gain of 1.74\%. This suggests that domains with higher linguistic variability and conversational structure benefit significantly from domain-adaptive pretraining, as the MLM step allows the model to internalize subtitle-specific vocabulary and expression patterns.

The MR (Movie Reviews) dataset also shows a notable improvement of 0.98\%, indicating that contextual understanding within formal review text is effectively enhanced by the proposed approach.

The GT (Generic Tweets) dataset shows a moderate improvement of 0.40\%, reflecting the benefit of back-translation augmentation in handling informal and noisy text such as social media content. In contrast, the PT (Political Tweets) dataset exhibits a relatively smaller gain of 0.12\%, likely because its strong baseline performance leaves limited room for further improvement.

\begin{figure*}[t]
\centering
\includegraphics[width=0.85\textwidth]{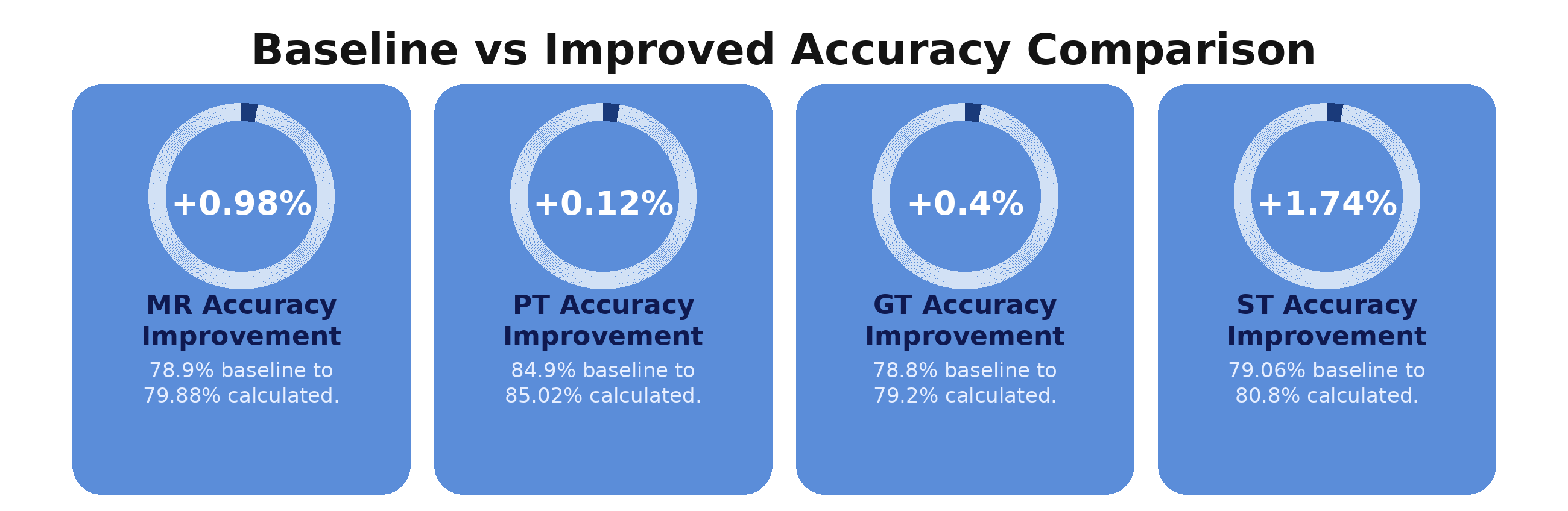}
\caption{Comparison of baseline and improved accuracies across different domains.}
\label{fig:results}
\end{figure*}

Overall, the findings show that the proposed methodology provides consistent and reliable improvements across all domains. The combination of data augmentation and domain-adaptive pretraining helps the model better handle linguistic diversity and domain-specific characteristics, making it more robust for real-world sentiment analysis applications. These results confirm that domain-adaptive pretraining and data augmentation are valuable techniques for improving sentiment analysis in low-resource languages.

\section{Conclusion}

In this paper, a transformer-based method for domain-specific sentiment classification in Marathi utilizing the MahaBERT model is presented. The approach addresses challenges of low-resource languages and domain variability by integrating data augmentation and domain-adaptive pretraining within a unified framework.

Experimental results show consistent performance improvements across all domains of the MahaSent-MD dataset. The observed gains, ranging from 0.12\% to 1.74\%, indicate that combining back-translation augmentation with masked language model pretraining effectively enhances the model's ability to capture domain-specific linguistic patterns.

Larger gains are observed in domains with greater linguistic variability, while already strong-performing domains show smaller improvements. This highlights the approach's effectiveness in adapting to domain-specific characteristics and producing more robust predictions. Overall, incorporating diverse training data leads to more stable and reliable performance across different text types.

Overall, the methodology offers a reliable solution for Marathi sentiment analysis that can be extended to other low-resource Indic languages. Future work includes exploring larger transformer architectures, multilingual transfer learning, and more sophisticated augmentation methods to further improve cross-domain performance.

\section*{Acknowledgments}

Under the L3Cube Pune mentorship program, this work was completed. We would like to sincerely thank our mentors at L3Cube for their unwavering encouragement and support during this endeavor. This work was carried out under L3Cube-MahaNLP Project \cite{joshi2022mahanp}.

\bibliography{main}

\end{document}